\documentclass[pdflatex,sn-mathphys]{sn-jnl}

\jyear{2023}%

\theoremstyle{thmstyleone}%
\theoremstyle{thmstyletwo}%

\theoremstyle{thmstylethree}%
\newtheorem{definition}{Definition}%

\usepackage{subcaption}

\begin{document}

\title[Interpretable Machine Learning for Pregnancy]{Interpretable Predictive Models to Understand Risk Factors for Maternal and Fetal Outcomes}




\author*[1]{\fnm{Tomas M.} \sur{Bosschieter}}\email{tomasbos@stanford.edu}
\author[1]{\fnm{Zifei} \sur{Xu}}
\author[1]{\fnm{Hui} \sur{Lan}}
\author[2]{\fnm{Benjamin J.} \sur{Lengerich}}
\author[3]{\fnm{Harsha} \sur{Nori}}
\author[4]{\fnm{Ian} \sur{Painter}}
\author*[4]{\fnm{Vivienne} \sur{{Souter~MD}}} \email{vsouter@qualityhealth.org}
\author*[3]{\fnm{Rich} \sur{Caruana}}\email{rcaruana@microsoft.com}
\affil[1]{\orgname{Stanford University}, \orgaddress{\city{Stanford}, \country{USA}}}
\affil[2]{\orgname{Massachusetts Institute of Technology}, \orgaddress{\city{Cambridge}, \country{USA}}}
\affil[3]{\orgname{Microsoft Research}, \orgaddress{\city{Redmond}, \country{USA}}}
\affil[4]{\orgname{Foundation for Healthcare Quality}, \orgaddress{\city{Seattle}, \country{USA}}}






\abstract{
\textbf{Purpose:} 
Although most pregnancies result in a good outcome, complications are not uncommon and can be associated with serious implications for mothers and babies. Predictive modeling has the potential to improve outcomes through better understanding of risk factors, heightened surveillance for high risk patients, and more timely and appropriate interventions, thereby helping obstetricians deliver better care. 

We identify and study the most important risk factors for four types of pregnancy complications: (i) severe maternal morbidity, (ii) shoulder dystocia, (iii) preterm preeclampsia, and (iv) antepartum stillbirth. 

\textbf{Methods:} We use an Explainable Boosting Machine (EBM), a high-accuracy glass-box learning method, for prediction and identification of important risk factors. We undertake external validation and perform an extensive robustness analysis of the EBM models.

\textbf{Results:} EBMs match the accuracy of other black-box ML methods such as deep neural networks and random forests, and outperform logistic regression, while being more interpretable. EBMs prove to be robust. 

\textbf{Conclusion:} The interpretability of the EBM models reveals surprising insights into the features contributing to risk (e.g. maternal height is the second most important feature for shoulder dystocia) and may have potential for clinical application in the prediction and prevention  of serious complications in pregnancy.

}




\maketitle

\section*{Acknowledgments}
This preprint has not undergone peer review or any post-submission improvements or corrections. The Version of Record of this article is published in the Journal of Healthcare Informatics Research, and is available online at \url{http://dx.doi.org/10.1007/s41666-023-00151-4}. As part of the Springer Nature Content Sharing Initiative, we are able to share full-text access to a view-only version of the paper through the following SharedIt link:\\ \url{https://rdcu.be/dowFT}. We thank Springer Nature for their commitment to content sharing and the SharedIt initiative for facilitating this.

We are indebted to Kristin Sitcov, Executive Director Clinical Programs at the Foundation for Health Care Quality (FCHQ), for her support for this project.

\section{Introduction}\label{sec:introduction}
Of the 3.6 million births per year in the U.S. \cite{osterman2022births}, Severe Maternal Morbidity (SMM) happens in as many as 60,000 cases \cite{declercq2020maternal}, leading to serious short- or long-term consequences for the mother's health.\footnote{In our work, SMM is a composite term for 6 adverse diagnoses: (i) hysterectomy, (ii) blood transfusion, (iii) disseminated intravascular coagulation, (iv) amniotic fluid embolism, (v) thromboembolism, and (vi) eclampsia.}  Shoulder dystocia and (preterm) preeclampsia are two other relatively common, serious conditions. Antepartum stillbirth is a relatively rare, but devastating outcome. If accurate models could be trained that were interpretable and provided timely risk prediction, obstetric providers could provide more personalised care, potentially averting some cases of SMM, shoulder dystocia, preterm preeclampsia and antepartum stillbirth.  \cite{creanga2014racial, declercq2020maternal}. In this work we use glassbox Explainable Boosting Machines (EBMs) to train interpretable, high accuracy machine learning models for each of these outcomes.

When modeling risk for SMM, preterm preeclampsia and antepartum stillbirth, we only use features known early in pregnancy.  For shoulder dystocia we use all features known just prior to delivery. While several analyses and (black-box) models have been developed to predict SMM and its risk factors \cite{gao2019learning, lengerich20211017, cartus2021can, callaghan2008identification, lengerich202146}, (preterm) preeclampsia \cite{bennett2022imbalance, moreira2017predicting, bosschieter2023preterm, jhee2019prediction}, shoulder dystocia \cite{tsur2020development, bartal2021651}, and antepartum stillbirth \cite{bosschieter2023unique, allotey2022external, malacova2020stillbirth}, the risk factors for each have remained under-studied using interpretable machine learning. Additionally, few models are externally validated, i.e. validated on a hold-out set containing data from a disjoint set of hospitals \cite{steyerberg2016prediction, wynants2017key, kleinrouweler2016prognostic}. Furthermore, models trained on international data (e.g. \cite{squires2011us}) might not perform as well in U.S.-based hospitals \cite{yamada2021external}.

Equipped with recent clinical data containing 158,629 de-identified birth events and over 100 features, we use interpretable machine learning models, Explainable Boosting Machines (EBMs), to uncover the most important risk factors for each outcome. We show that EBMs yield an Area Under the Receiver Operating Characteristic curve (AUROC) on par with XGBoost \cite{chen2016xgboost}, random forests \cite{breiman2001random}, and deep neural networks (DNNs) \cite{gardner1998artificial}, while outperforming logistic regression \cite{kleinbaum2002logistic}. Additionally, we provide empirical evidence that EBMs are robust by showing they generalize well and yield well-calibrated probabilities, and by evaluating the shape functions of the EBM models as they evolve as a function of the training data set size. We compute the discrete Fr\'echet distances from the shape functions trained on subsets of the data to the shape function trained on the largest data set, which shows the convergence of the sequence of shape functions. Robustness of models is an especially important consideration in healthcare \cite{qayyum2020secure, steyerberg2016prediction}.

Our main contributions are:
\begin{itemize}
     \item Illustrating the value of interpretable models in four healthcare case studies.   
     \item Leveraging a new and robust data set in healthcare to model risk of SMM, shoulder dystocia, preterm preeclampsia, and antepartum stillbirth.
     \item Demonstrating how intelligible models reveal surprising risk factors not traditionally recognized as important.
    \item Empirically evaluating the robustness, accuracy and calibration of EBMs.
\end{itemize}

\section{Preliminaries}\label{sec:preliminaries}

\subsection{Explainable Boosting Machines}\label{sec:EBM}
\begin{figure}[!h]
    \centering
    \includegraphics[width=.9\linewidth]{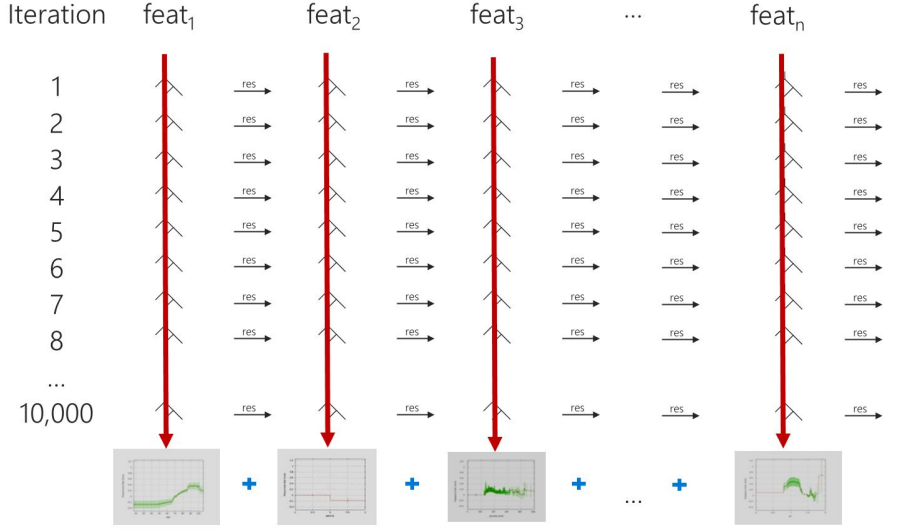}
    \caption{EBMs are GAMs that perform cyclic gradient boosting with shallow decision trees. As an additive models, the set of trees trained on a feature could be ensembled to form the shape function of that feature. This image illustrates the ensembling (with the red vertical arrows) as well as the training process more generally. Details can be found in \cite{nori2019interpretml}.}
    \label{fig:EBM_training}
\end{figure}

For a target variable $Y$ and predictor features $x_1, \ldots, x_n$, Generalized Additive Models (GAMs) \cite{hastie2017generalized} generalize linear (regression) models $Y = b_0 + a_1x_1 + \cdots + a_nx_n$ to an additive model with univariate shape functions $f_i$ and a situation-dependent link function $g$ (e.g. logistic for classification and identity for regression):
\begin{equation}\label{eq:GAM}
    g(\mathbb{E}[Y]) = \beta_0 + f_1(x_1) + f_2(x_2) + \cdots + f_n(x_n).
\end{equation}
The Explainable Boosting Machine (EBM) is an algorithm for training the $f_i$ by boosting shallow decision trees in a round-robin fashion \cite{lou2012intelligible, nori2019interpretml} (see Figure~\ref{fig:EBM_training}). Additionally, EBMs can be trained to automatically detect important pairwise interaction terms to further boost accuracy while preserving intelligibility \cite{lou2013accurate}. Because EBMs are GAMs, the trees trained on the $i$-th feature, $x_i$, combine to provide a component function $f_i(x_i)$ which is piecewise constant. 
Plotting $f_i(x_i)$ provides the end-user with a visual representation of the function learned by the tree ensemble, e.g. those seen at the bottom of Figure~\ref{fig:EBM_training}. Observe that through this training process and having a main effect $f_i(x_i)$ for each feature $x_i$, each feature's shape function is learned in parallel with and corrected for all other features in the model. This additivity is an important property for intelligibility and transparency.

EBM models use a combination of inner- and outer-bagging. At each iteration and for each feature $x_i$, multiple subsets are taken from the residual, and an ensemble is constructed from all shallow trees trained on the samples. This is called inner-bagging.  Outer-bagging refers to bagging the entire process, i.e. retraining the entire model multiple times on random subsamples of the data. This combination of inner- and outer-bagging allows EBM models to improve accuracy, reduce variance, make the learned $f_i(x_i)$ smoother and easier to interpret, and also provide confidence intervals on the $f_i(x_i)$.

EBMs have recently gained popularity because of their high accuracy and local and global interpretability, with applications ranging from healthcare \cite{zhou2020identifying} to predicting sports outcomes \cite{decroos2019interpretable, xenopoulos2022analyzing}, slope failures \cite{maxwell2021explainable}, and modeling dark matter \cite{cannarozzo2021merger}.

\subsection{Discrete Fr\'echet distance}\label{sec:Frechet}
To evaluate the robustness of EBMs we examine how the shape functions of EBMs converge as the number of samples in a dataset increases. 
Since the models might find different best-fit splits in the feature values when training decision trees for different data sets, the shape functions might learn different bins for each data set size. This makes it hard to compare shape functions using standard measures such as the root-mean-square error (RMSE). Instead, we turn to the discrete Fr\'echet distance for polygonal chains, a measure of how similar two curves are given that the partitions in each might be different. The Fr\'echet distance is defined as follows:

\begin{definition}[Fr\'echet distance \cite{alt1995computing}] \label{def:frechet-distance}
Let $(S,d)$ be a metric space and $A, B \colon [0,1] \to S$ curves in $S$, i.e. continuous mappings. Denoting $\alpha, \beta \colon [0,1] \to [0,1]$ reparameterizations, the Fr\'echet distance between A and B is defined as
\begin{equation}
    \text{d}_F(A,B) = \inf_{\alpha, \beta} \max_{t \in [0,1]} \left\{ d(A(\alpha(t)), B(\beta(t))) \right\}.
\end{equation}
\end{definition}

\section{Methods}\label{sec:methods}

\subsection{Data set}\label{sec:dataset}
The clinical data used in this analysis are from the Foundation for Health Care Quality’s Obstetrical Care Outcomes Assessment Program (OB COAP) \cite{fhcq_obcoap}, including de-identified patient-level information from medical records from January 2016 till December 2021 at 20 hospitals from Northwestern U.S. collected for quality improvement purposes. The data set contains 158,629 births, but specific filters are applied per outcome, lowering the number of samples in the data set used for each outcome.

For shoulder dystocia we consider only singleton births and no cesarean births, while excluding antepartum stillbirths, so that we consider only the population for whom shoulder dystocia risk prediction is useful. For antepartum stillbirth, we restrict the population to singletons to remove ambiguity about which fetus the outcome is for. The number of samples, positive samples, and features used for each outcome are listed in Table~\ref{tab:datasets}. See Section~\ref{sec:feature_engineering-preproc_assump} for details on filtering and preprocessing of the initial data set as a whole.


\begin{table}[!h]
    \centering
    \caption{Data set characteristics per outcome.}
    \begin{tabular}{c|c|c|c}
        \textbf{Outcome} & \textbf{Num. samples} & \textbf{Num. positives} & \textbf{Num. features}  \\
        \hline
         SMM & 155,935 & 2262 (1.5\%) & 49 \\
         Shoulder dystocia & 82,889 & 2794 (3.4\%) & 56 \\
         Preterm preeclampsia & 153,432 & 2919 (1.9\%) & 29 \\
         Antepartum stillbirth & 153,432 & 610 (0.4\%) & 29
    \end{tabular}
    \label{tab:datasets}
\end{table}

\subsubsection{Feature engineering and preprocessing assumptions} \label{sec:feature_engineering-preproc_assump}
For each outcome we use clinical expertise to identify and select the appropriate features, i.e. those that can be measured before the outcome occurs (with one exception: the shoulder dystocia model includes birthweight, a feature that can only be estimated before birth). For example, in trying to predict preterm preeclampsia (i.e. preeclampsia before 37 weeks' gestation)\footnote{Note that predicting preterm preeclampsia can only be done in practice $< 37$ weeks into the pregnancy per definition of preterm.}, a feature such as `time from hospital admission to delivery' should not be used by a machine learning model in our setting. This is because labor has not taken place yet and so this feature's value is not known at the time of prediction, i.e. $< 37$ weeks' gestation. However, the feature ‘time from hospital admission to delivery’ in the data set is selected as a feature to be used in predicting shoulder dystocia because it's known at delivery. Full lists of included features, ranked by importance to each risk prediction outcome, can be found in Appendix~\ref{sec_App:List_features_rankings}.

As a preprocessing step, we discard birth events with data deemed implausible or impossible by clinical experts (and assumed to be entered into the system erroneously), taking a conservative approach to avoid accidentally discarding valid data. For example, we never include births with a negative length of time from admission till delivery, births where the baby’s birthweight is over 8000 grams, and cases where the pregnant patient has a BMI over 120 $\text{kg}/\text{m}^2$. Furthermore, we use the standard procedure of dummy encoding for categorical data, and impute missing values for continuous features with the mean and missing values for categorical features with a unique identifier, typically $-1$. Data normalization is not required for tree-based methods such as EBMs, random forests and boosted trees because trees are scale-invariant.


\subsection{Experimental setup}\label{sec:experimental-setup}
To obtain training and external validation data sets, we first stratify births by hospital. The set of hospitals $\mathcal{H}$ is then partitioned into 3 sets according to the level of care they provide. There are 7 level-1 hospitals, 6 level-2 hospitals, and 7 level-3 hospitals, each hospital containing a different number of samples. To do external validation, we partition the hospitals $\mathcal{H}$ into 2 sets, where the births in the first set of hospitals constitutes the training set, and the second set of hospitals contains the ``external validation" set (consisting of births from a disjoint set of hospitals).  To ensure (near) equal representation of levels of care in each of the training and external validation sets, we generate many two-way partitions of the hospitals, and select only those which at the aggregate have: 1) a near 75\%/25\% split between train and test patients, and 2) yield more than 1 hospital of each level in each set so as to minimize the risk of overfitting to any one hospital and level.

The patients in the 25\% set are from a disjoint set of hospitals that are used for external validation. External validation is important in healthcare to ensure that the learned models are robust, generalize well to other hospitals, and are well-calibrated for patients at all risk levels.  We use a bootstrap analysis to generate confidence intervals for the AUCs reported in Table~\ref{tab:aurocs}.


\subsubsection{Model parameters}\label{sec:model-parameters}
The hyperparameters of all models were each determined using 5-fold randomized search Cross-Validation to maximize AUROCs while ensuring good calibration; all parameters not mentioned are common defaults. For SMM, shoulder dystocia and preterm preeclampsia: for EBMs, we use hyperparameters \verb|outer_bags=25|, \verb|inner_bags=25|, \verb|min_samples_leaf=25|, \verb|interactions=10|. For XGBoost we use \verb|eta=0.04|, \verb|subsample=0.7|, and \verb|max_depth=5|. Random forests are trained with \verb|n_estimators=1000|, \verb|min_samples_split=60|, and \verb|min_samples_leaf=40|. Lastly, the DNN is a multilayer preceptron (MLP) with 7 hidden layers of 100 neurons each. For antepartum stillbirth, the outcome with the smallest number of positives, optimization lowers the number of \verb|min_samples_leaf| to $15$ for EBMs, and $25$ for random forests.

\subsection{Evaluating EBM robustness}\label{sec:robustness-EBMs-methods}
While the robustness of many models has been well-studied, it appears this is not fully the case yet for Explainable Boosting Machines. We provide empirical evidence that EBMs are robust in that its constituent shape functions converge rapidly as a function of data sample size and the number of positives therein, and that they generalize well. The latter is shown by performing external validation, see Section~\ref{sec:results}. The former is done by showing (i) the shape functions' evolutions visually (see Section~\ref{sec:robustness-EBMs-results}), and (ii) the sequence of Fr\'echet distances of the continuous shape functions trained on subsampled data sets compared to the final shape function trained on all data. For discrete features, we use RMSE instead of Fr\'echet distance.

For each of the 4 outcomes we select one important feature to perform this analysis on. We examine 2 continuous features and 2 categorical features, one of which is binary. Specifically, the outcome-feature pairs are:
\begin{itemize}
    \item \textbf{Shoulder dystocia}: \textit{birthweight} (continuous). Birthweight is the most important predictor for shoulder dystocia. See Figure~\ref{fig:ShDys-shapefunc-evolution}.
    \item \textbf{Preterm preeclampsia}: \textit{maternal BMI} (continuous). Maternal BMI is chosen because of its complex shape function: smooth and near-linear in some regions, while also exhibiting significant jumps in other regions, presumably due to clinical interventions. See Figure~\ref{fig:PP-shapefunc-evolution}.
    \item \textbf{SMM}: \textit{nulliparity} (binary). Nulliparity is the second most important feature for SMM, and traditionally considered a crucial risk indicator. See Figure~\ref{fig:SMM-shapefunc-evolution}.
    \item \textbf{Antepartum stillbirth}: \textit{race} (categorical). Race is the second most important feature in predicting antepartum stillbirth, and contains multiple classes. See Figure~\ref{fig:SB-evolution_shapefuncs}. 
\end{itemize}


\section{Results}\label{sec:results}
For each outcome (SMM, shoulder dystocia, preterm preeclampsia, antepartum stillbirth) we compare the EBM's performance to those of XGBoost, random forests, deep neural networks, and logistic regression. The mean Area Under the ROC curve (AUROC) for each model and outcome can be found in Table~\ref{tab:aurocs}. All results presented are obtained through external validation.

\begin{table*}[!h]
    \caption{A comparison between the Area Under the ROC curve (AUROC) for each outcome. Higher is better. The EBM has the highest average AUROC.}
    \centering

    \scalebox{0.65}{
    \begin{tabular}{c|c c c c c}
    \toprule
          Outcome & \textbf{EBM} & \textbf{Logistic Regression} & \textbf{XGBoost} & \textbf{Random Forests} & \textbf{DNN} \\
         \midrule
         SMM & $ 0.700 \pm 0.013$ & $0.683 \pm 0.015$ & $0.679 \pm 0.013$ & $0.683 \pm 0.015$ & $0.677 \pm 0.018$ \\
         Shoulder dystocia  & $0.744 \pm 0.017$ & $0.742 \pm 0.015$ & $0.751 \pm 0.018$ & $0.713 \pm 0.019$ & $0.752 \pm 0.019$ \\
         Preterm preeclampsia & $0.767 \pm 0.013$ &$0.735 \pm 0.010$ & $0.767 \pm 0.010$ & $0.749 \pm 0.015$ & $0.754 \pm 0.014$ \\
         Antepartum stillbirth  &$0.710 \pm 0.011$ & $0.691 \pm 0.010$ & $0.714 \pm 0.012$ & $0.713 \pm 0.017$ & $0.706 \pm 0.019$ \\
         \midrule
         Mean AUROC & $\mathbf{0.730 \pm 0.014}$ & $0.713 \pm 0.013$ & $\mathbf{0.728 \pm 0.013}$ & $0.715 \pm 0.017$ & $0.722 \pm 0.018$
    \end{tabular}
    }
    \label{tab:aurocs}
\end{table*}

EBMs allow each feature's contribution to the log-odds of the risk to be easily visualized since EBMs are additive models, as discussed in Section~\ref{sec:EBM}. For each outcome we present 2 shape functions together with a feature importance plot, which shows the top 15 features in terms of their feature importance, i.e. their mean absolute contribution to the log-odds of the outcome prediction. This is more informative than simply a ranking of features, as given in Appendix~\ref{sec_App:List_features_rankings}, as the actual measured feature importance is reported in the figure. Note that for the shape functions, the error bars shown represent the standard deviations yielded by the outer-bagging process employed by EBMs.

\subsubsection{SMM} \label{sec:results-SMM}
\begin{figure*}[!h]
    \centering
    \begin{subfigure}[t]{.4\textwidth}
        \centering
        \includegraphics[width=1\linewidth]{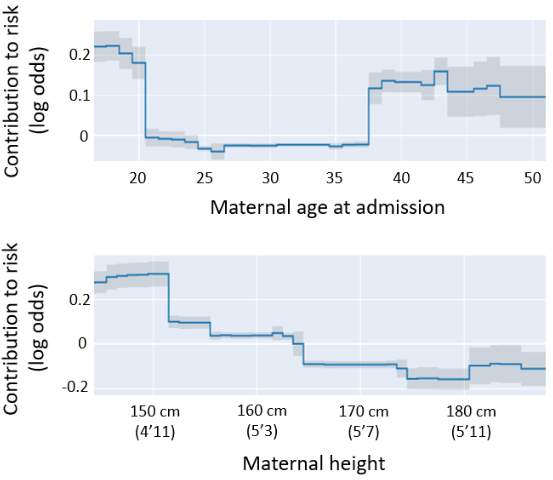}
        \caption{Risk contributions.}
        \label{fig:SMM-shapefunctions}
    \end{subfigure}%
    \hfill
    \begin{subfigure}[t]{.56\textwidth}
        \centering
        \includegraphics[width=1\linewidth]{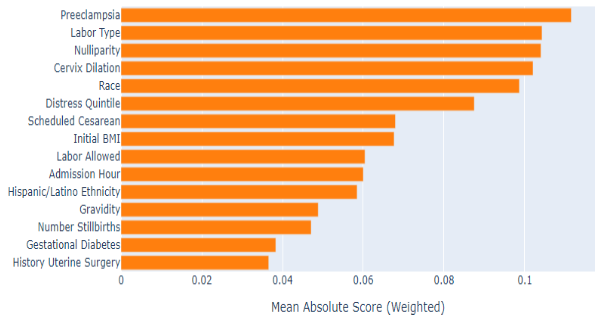}
        \caption{Feature importance plot of SMM.}
        \label{fig:SMM-ft_imp}
    \end{subfigure}%
    \caption{The shape functions learned by the EBM for SMM of (i) maternal age at admission, and (ii) maternal height, along with SMM's feature importance plot \cite{xu2023predicting}. The full list of ranked features can be found in Appendix~\ref{sec_App:ft_ranking_SMM}.}
\label{fig:SMM_results}
\end{figure*}

The shape functions of the \textit{maternal age at admission} and \textit{maternal height} features are in Figure~\ref{fig:SMM-shapefunctions}, while the feature importance plot is in Figure~\ref{fig:SMM-ft_imp}. The most important features according to the EBM are \textit{preeclampsia/gestational hypertension}, \textit{labor type}\footnote{Labor type refers to whether labor was spontaneous or induced.}, \textit{nulliparity}, \textit{cervical dilation}, \textit{race}, and the \textit{Distressed Communities Index (DCI) quintile} \cite{DCIwebsite}. The complete, ranked list of features used to train all models for SMM risk prediction can be found in Appendix~\ref{sec_App:ft_ranking_SMM}. 

\subsubsection{Shoulder dystocia} \label{sec:results-ShDys}
\begin{figure*}[!h]
    \centering
    \begin{subfigure}[t]{.4\textwidth}
        \centering
        \includegraphics[width=1\linewidth]{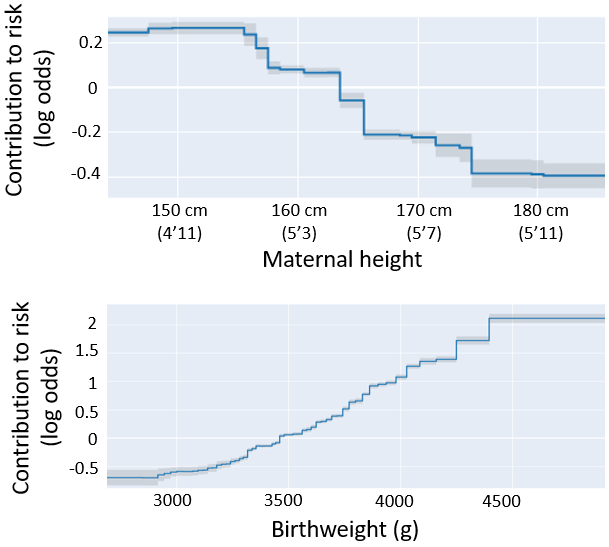}
        \caption{The individual contributions of baby weight and maternal height to shoulder dystocia risk.}
        \label{fig:ShD-babyweight_and_maternal_height}
    \end{subfigure}%
    \hfill
    \begin{subfigure}[t]{.56\textwidth}
        \centering
        \includegraphics[width=1\linewidth]{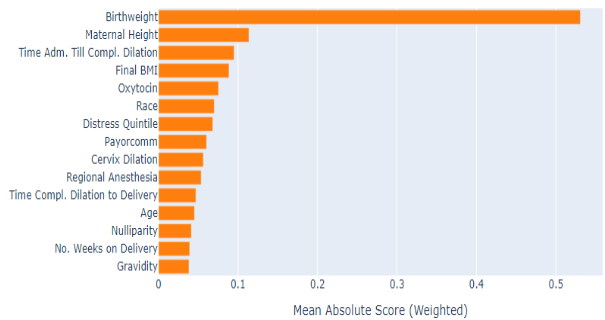}
        \caption{The feature importance plot for shoulder dystocia. Full list of ranked features can be found in Appendix~\ref{sec_App:ft_ranking_ShDys}.}
        \label{fig:ShD-ft_imp}
    \end{subfigure}%
    \caption{Two shape functions of the EBM for shoulder dystocia, as well as the feature importance plot. \cite{lan2023understanding}}
\label{fig:ShD_results}
\end{figure*}

The EBM's shape functions for \textit{maternal height} and the \textit{baby's birthweight} can be found in Figure~\ref{fig:ShD-babyweight_and_maternal_height}, while Figure~\ref{fig:ShD-ft_imp} shows the feature importance plot. The complete list of importance-ranked features can be found in Appendix~\ref{sec_App:ft_ranking_ShDys}. 

We observe a clinically surprising near-linear (in log odds) relationship between \textit{birthweight} and risk in Figure~\ref{fig:ShD-babyweight_and_maternal_height} for weights between 3250g and 4250g. \textit{Birthweight} is clearly the most important feature in the feature importance ranking in Figure~\ref{fig:ShD-ft_imp}. The second most important feature is \textit{maternal height}, a feature not typically considered a major risk factor for shoulder dystocia. One might speculate that taller women might have larger pelvic openings on average than smaller women, resulting in a lower risk of shoulder dystocia. However, there are other factors that correlate with height, such as socioeconomic background.

\subsubsection{Preterm preeclampsia} \label{sec:results-PP}
\begin{figure*}[!h]
    \centering
    \begin{subfigure}[t]{.4\textwidth}
        \centering
        \includegraphics[width=1\linewidth]{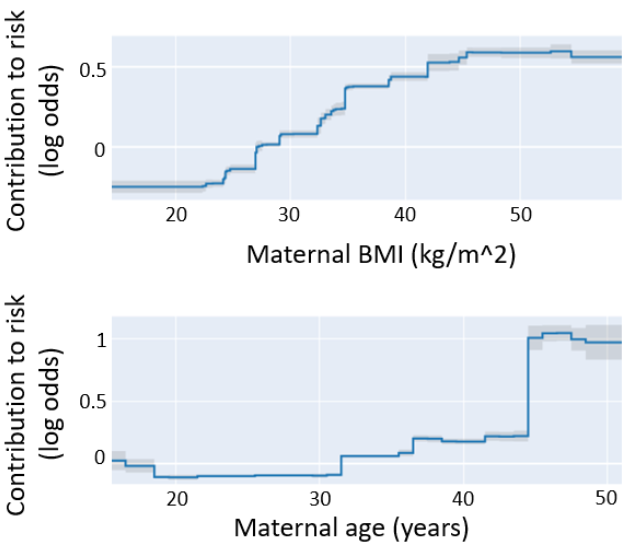}
        \caption{The risk contributions of maternal BMI and age to preterm preeclampsia. \cite{bosschieter2023preterm}}
        \label{fig:PP-mothersAge_BMI}
    \end{subfigure}%
    \hfill
    \begin{subfigure}[t]{.56\textwidth}
        \centering
        \includegraphics[width=1\linewidth]{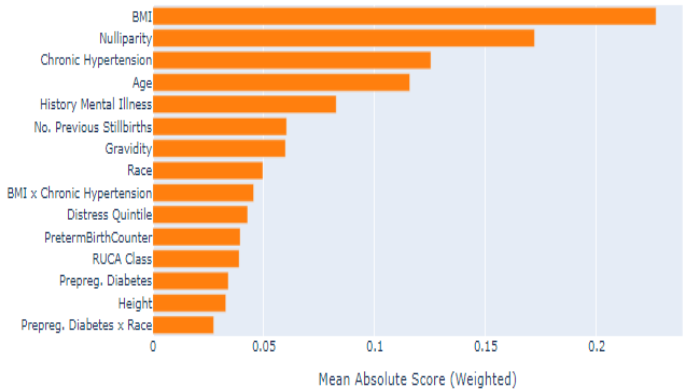}
        \caption{Feature importance plot for preterm preeclampsia risk. A full list of features can be found in Appendix~\ref{sec_App:ft_ranking_PP}.}
        \label{fig:PP-ft_imp}
    \end{subfigure}
    \caption{Shape functions for preterm preeclampsia along with the most important features.}
\label{fig:PP_results}
\end{figure*}

The EBM's shape functions of \textit{maternal BMI} and \textit{age} are shown in Figure~\ref{fig:PP-mothersAge_BMI}, while the feature importance plot is displayed in Figure~\ref{fig:PP-ft_imp}. A complete list of features used by the EBM can be found in Appendix~\ref{sec_App:ft_ranking_PP}, ranked by importance. We find that the most important risk factors for preterm preeclampsia are \textit{maternal BMI}, \textit{nulliparity}, \textit{chronic hypertension}, and \textit{age}. 

\subsubsection{Antepartum stillbirth} \label{sec:results-SB}
\begin{figure*}[!h]
    \centering
    \begin{subfigure}[t]{.4\textwidth}
        \centering
        \includegraphics[width=1\linewidth]{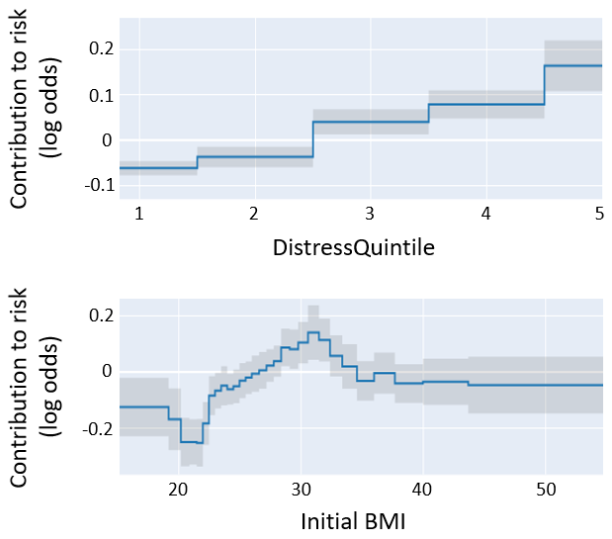}
        \caption{EBM's shape functions of the Distressed Communities Index (DCI) quintile and maternal BMI.}
        \label{fig:SB-shapefuncs}
    \end{subfigure}%
    \hfill
    \begin{subfigure}[t]{.56\textwidth}
        \centering
        \includegraphics[width=1\linewidth]{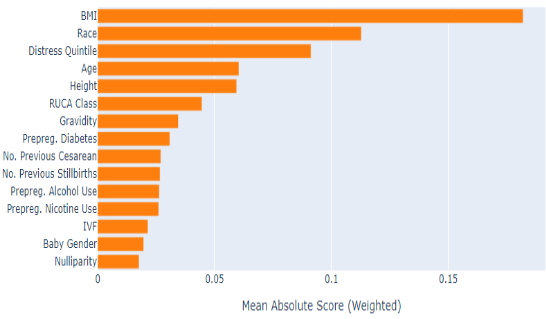}
        \caption{The feature importance ranking for antepartum stillbirth. Full list of ranked features can be found in Appendix~\ref{sec_App:ft_ranking_SB}.}
        \label{fig:SB-ft_imp}
    \end{subfigure}%
    \caption{The relationship between antepartum stillbirth risk and (i) the DCI quintile, and (ii) maternal BMI measured at the initial visit to the mother's healthcare provider. Also shown is the top 15 features ranked according to their feature importance. \cite{bosschieter2023unique}}
\label{fig:SB_images}
\end{figure*}

Antepartum stillbirth's most important features are listed in Figure~\ref{fig:SB-ft_imp}, while the shape functions for the \textit{Distressed Communities Index (DCI) quintile} and \textit{maternal BMI} are shown in Figure~\ref{fig:SB-shapefuncs}. The most important features seem to be \textit{maternal BMI}, while \textit{race}, \textit{DCI quintile}, \textit{maternal age}, \textit{maternal height}, and \textit{Rural-Urban Commuting Area (RUCA) class} \cite{RUCAwebsite} also appear to play important roles. A complete list of features used to train the ML models can be found in Appendix~\ref{sec_App:ft_ranking_SB}, where the features are ranked by feature importance.

\subsubsection{Race and ethnicity} \label{sec:results-Race_n_Eth}
Including race in clinical models is controversial \cite{jones2001invited, bedoya2022framework, paulus2020predictably}. However, if race or ethnicity were removed as a feature in the model, this would not necessarily make the model unbiased, because of correlation between race and ethnicity and other features in the model \cite{chang2021interpretable}. One way to mitigate bias in an EBM model is to include race and ethnicity as features when the model is trained, but then flatten or zero out the shape functions for race and ethnicity before deployment. 

\begin{figure*}[!h]
    \centering
    \begin{subfigure}[t]{.48\textwidth}
        \centering
        \includegraphics[width=1\linewidth]{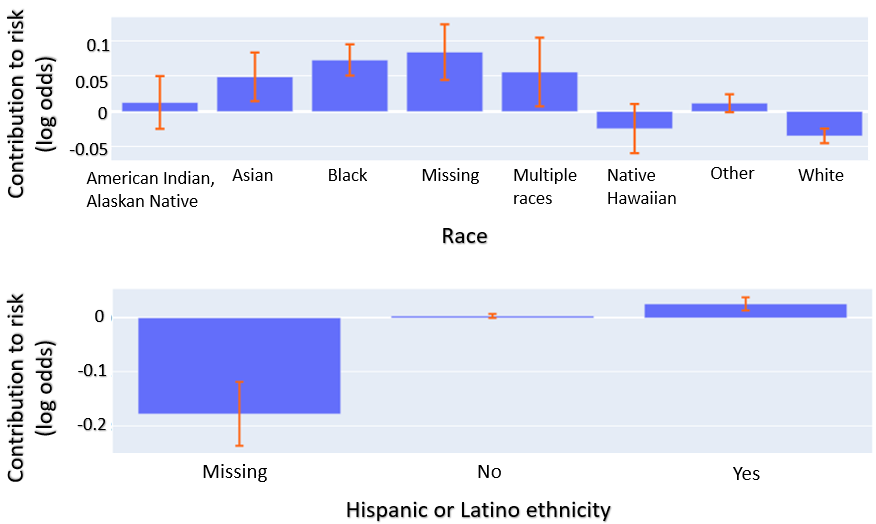}
        \caption{Shoulder dystocia.}
        \label{fig:ShD-race-ethn}
    \end{subfigure}%
    \hfill
    \begin{subfigure}[t]{.48\textwidth}
        \centering
        \includegraphics[width=1\linewidth]{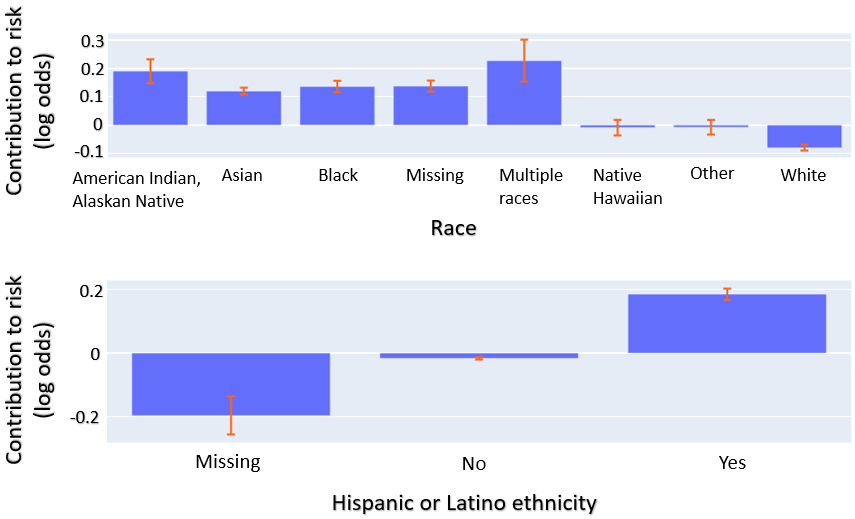}
        \caption{SMM.}
        \label{fig:SMM-race-ethn}
    \end{subfigure}
    \caption{The risk contribution of race and ethnicity for shoulder dystocia and SMM, after correcting for all other features. Preterm preeclampsia is similar. For more detail of race in antepartum stillbirth, see Section~\ref{sec:robustness_SB}. For SMM, the `Missing' bucket represents 5.6\% of the data for the feature `Race', and 3.7\% for `Hispanic or Latina ethnicity'. For shoulder dystocia, these are 4.8\% and 4.0\% respectively.}
\label{fig:race-ethn}
\end{figure*}

Figure~\ref{fig:race-ethn} shows the risk contribution of race and ethnicity for shoulder dystocia and SMM. The lowest risk for both shoulder dystocia and SMM seems to be in the group identified as White. The highest risk for shoulder dystocia is associated with pregnant people of missing, Black, and Asian race. Importantly, for SMM, the highest risk is associated with American Indian/Alaskan Native and Multiple races, even higher than for the Black pregnant population. It is noted however that the error bars are large enough for some race categories to overlap. For ethnicity, the lowest risk is associated with having missing data; the reason for this is unclear. Being of Hispanic/Latina ethnicity, when reported, is associated with a higher risk than non-Hispanic/Latina ethnicity.


\subsubsection{Model calibration} \label{sec:results-model_calibration}
\begin{figure*}[!h]
    \centering
    \begin{subfigure}[t]{.45\textwidth}
        \centering
        \includegraphics[width=1\linewidth]{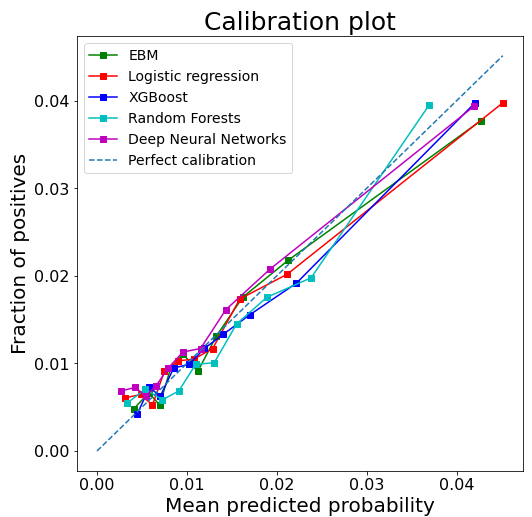}
        \caption{SMM.}
        \label{fig:calibration-SMM}
    \end{subfigure}%
    \hfill
    \begin{subfigure}[t]{.45\textwidth}
        \centering
        \includegraphics[width=1\linewidth]{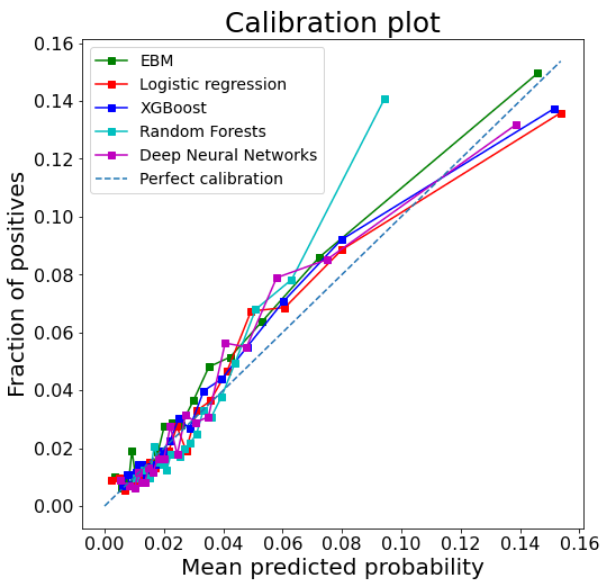}
        \caption{Shoulder dystocia.}
        \label{fig:calibration-ShDys}
    \end{subfigure}
    
    \medskip
    
    \begin{subfigure}[t]{.45\textwidth}
        \centering
        \includegraphics[width=1\linewidth]{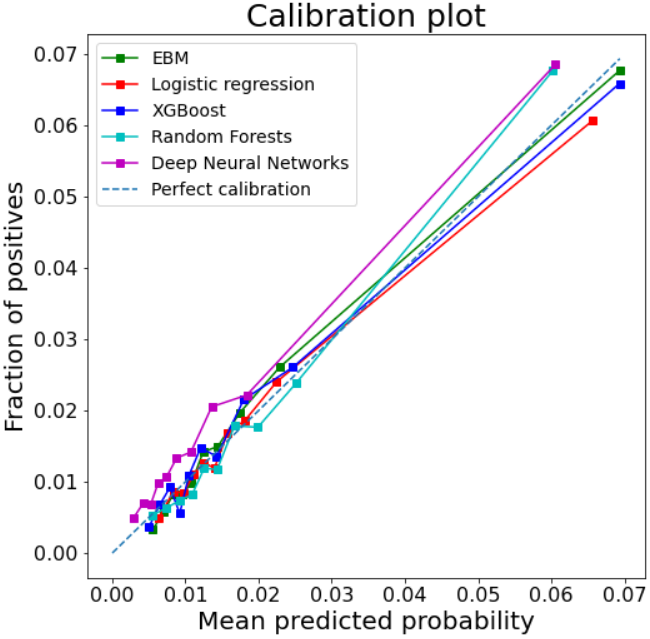}
        \caption{Preterm preeclampsia.}
        \label{fig:calibration-PP}
    \end{subfigure}%
    \hfill
    \begin{subfigure}[t]{.45\textwidth}
        \centering
        \includegraphics[width=1\linewidth]{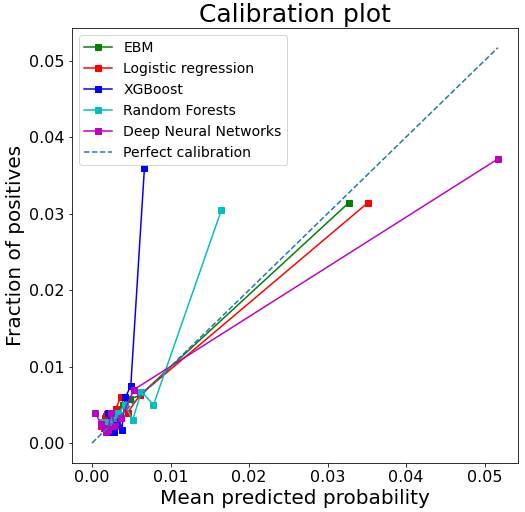}
        \caption{Antepartum stillbirth.}
        \label{fig:calibration-SB}
    \end{subfigure}
    \caption{Calibration plots of the EBM, logistic regression, random forests, XGBoost, and DNN models for SMM, shoulder dystocia, preterm preeclampsia, and antepartum stillbirth. 10 bins are used in the plots.}
\label{fig:cal-plots}
\end{figure*}

Another important consideration is model calibration. Because SMM, shoulder dystocia, preterm preeclampsia, and antepartum stillbirth are all unbalanced, ML models could be well-calibrated for the majority of patients, yet be uncalibrated for high-risk patients. This is why the within-level splits discussed in Section~\ref{sec:experimental-setup} are of such importance. The calibration plots, one for each outcome, are in Figure~\ref{fig:cal-plots}. All calibration plots are generated using 10 bins. Note that only EBM and logistic regression consistently yield good calibration, while XGBoost and Random Forests tend to slightly overfit and are overconfident, e.g. for antepartum stillbirth in Figure~\ref{fig:calibration-SB}, where bad calibration for the (top 10\%) highest-risk patients can be seen. Figure~\ref{fig:cal-plots} shows that EBMs do in fact attain calibration as good as or better than the gold standard, logistic regression. 

\subsection{Results of evaluating EBM robustness} \label{sec:robustness-EBMs-results}
We find that the EBMs are well-calibrated, and learn trends relatively well for individual features even when there are (very) few positives in the data set. Following our experimental setup as described in Section~\ref{sec:experimental-setup}, we show the evolution of the shape functions for 2 continuous and 2 categorical features: \textit{nulliparity} (categorical) for SMM, \textit{birthweight} (continuous) for shoulder dystocia, \textit{maternal BMI} (continuous) for preterm preeclampsia, and \textit{Race} (categorical) for antepartum stillbirth. We present the evolution of the shape functions for the continuous features first.

\subsubsection{Shoulder dystocia} \label{sec:robustness_ShDys}
Figure~\ref{fig:ShD-ft_imp} showed that birthweight is the most important feature in shoulder dystocia risk prediction, and we now demonstrate how that feature develops while trained with random subsets of sizes 500, 1000, 2000, 5000, 10000, 20000, 30000, 40000, 50000, 60000, 70000, and the entire data set containing 82889 birth events. 

\begin{figure}[!h]
    \centering
    \includegraphics[width=\linewidth]{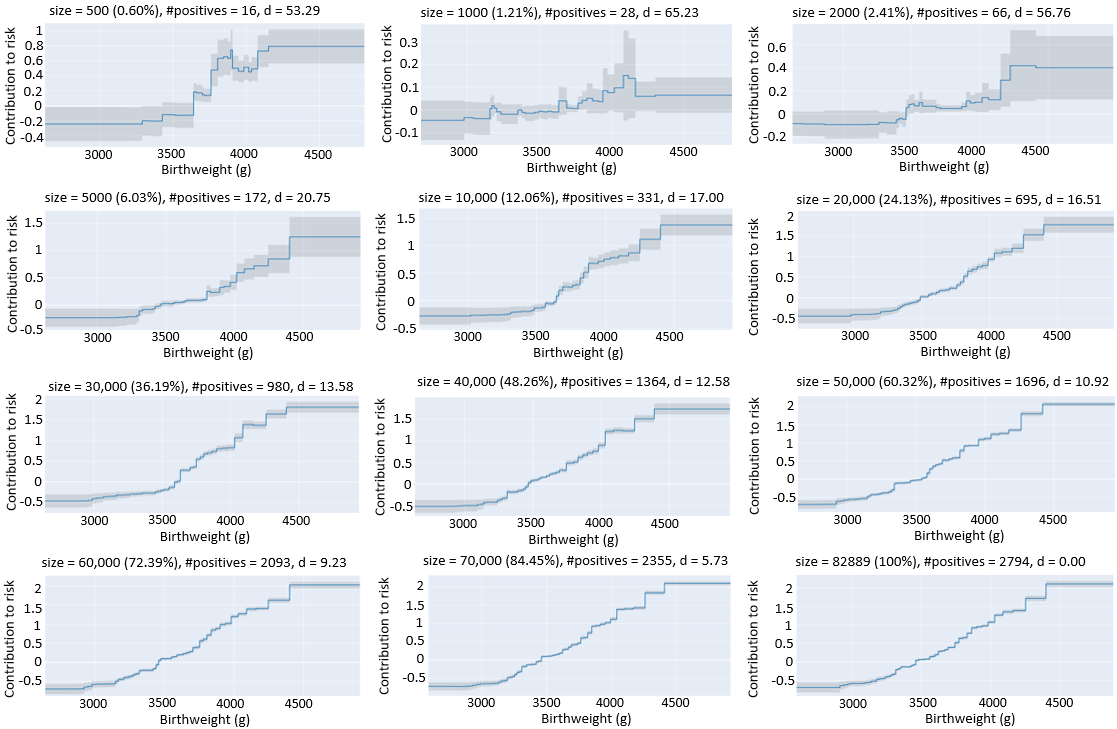}
    \caption{Evolution of the shape functions of \emph{birthweight} while predicting shoulder dystocia, as the data set size (``size" in the figure) is varied. The parameter ``d" represents the discrete Fr\'echet distance discussed in Section~\ref{sec:Frechet}.}
    \label{fig:ShDys-shapefunc-evolution}
\end{figure}

\subsubsection{Preterm preeclampsia} \label{sec:robustness_PP}
\begin{figure}[!h]
    \centering
    \includegraphics[width=\linewidth]{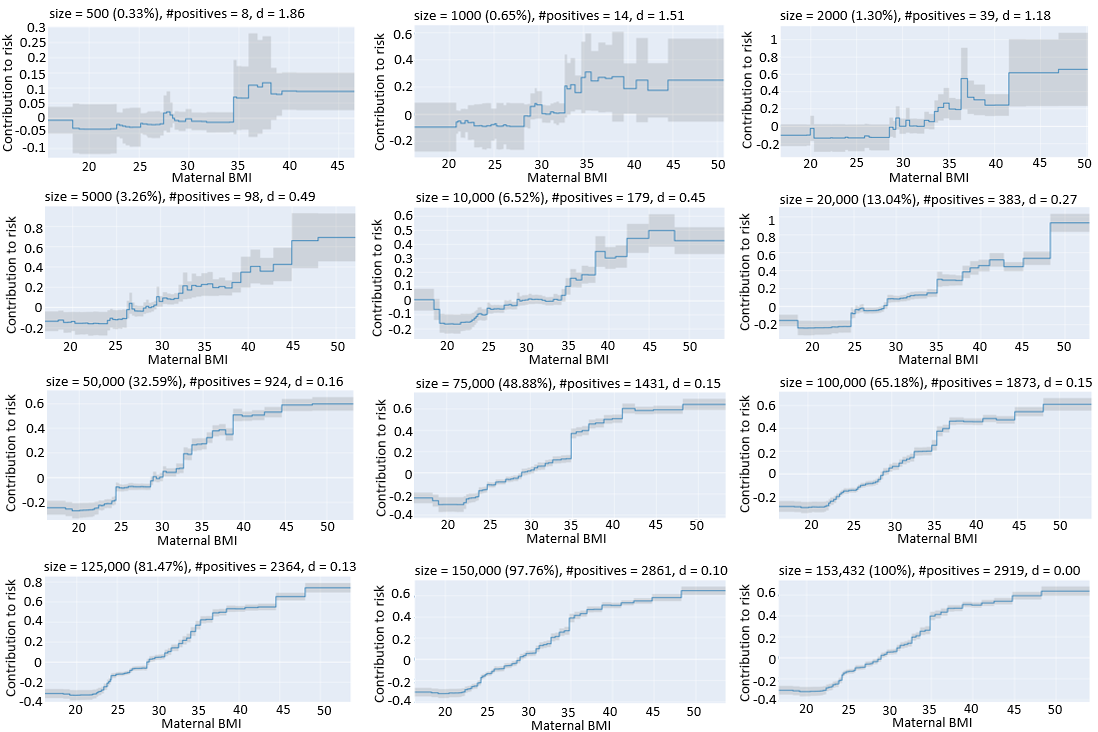}
    \caption{Evolution of the shape functions of maternal BMI while predicting preterm preeclampsia, as the data set size (``size" in the figure) is varied. The discrete Fr\'echet distance to the final shape function is reported as ``d".}
    \label{fig:PP-shapefunc-evolution}
\end{figure}

See Figure~\ref{fig:PP-shapefunc-evolution} for the evolution of the shape function of \emph{initial maternal BMI}, a continuous feature used to train the EBM in predicting preterm preeclampsia risk. We again report the size of the sampled data set and the number of positives therein, where we select data set sizes of 500, 1000, 2000, 5000, 10000, 20000, 50000, 75000, 100000, 125000, 150000, and the entire data set of size 153432. Additionally, ``d" represents the discrete Fr\'echet distance to the final shape function, representing its similarity to the best-fit shape function. We observe that the Fr\'echet distance decreases rapidly after a data set size of 2000 (1.30\%) with only 39 positives, see both Figure~\ref{fig:PP-shapefunc-evolution} for the shape function evolution, as well as the Fr\'echet distance graph in Figure~\ref{fig:frechetdists}.

\subsubsection{SMM} \label{sec:robustness_SMM}
\begin{figure}[!h]
    \centering
    \includegraphics[width=\linewidth]{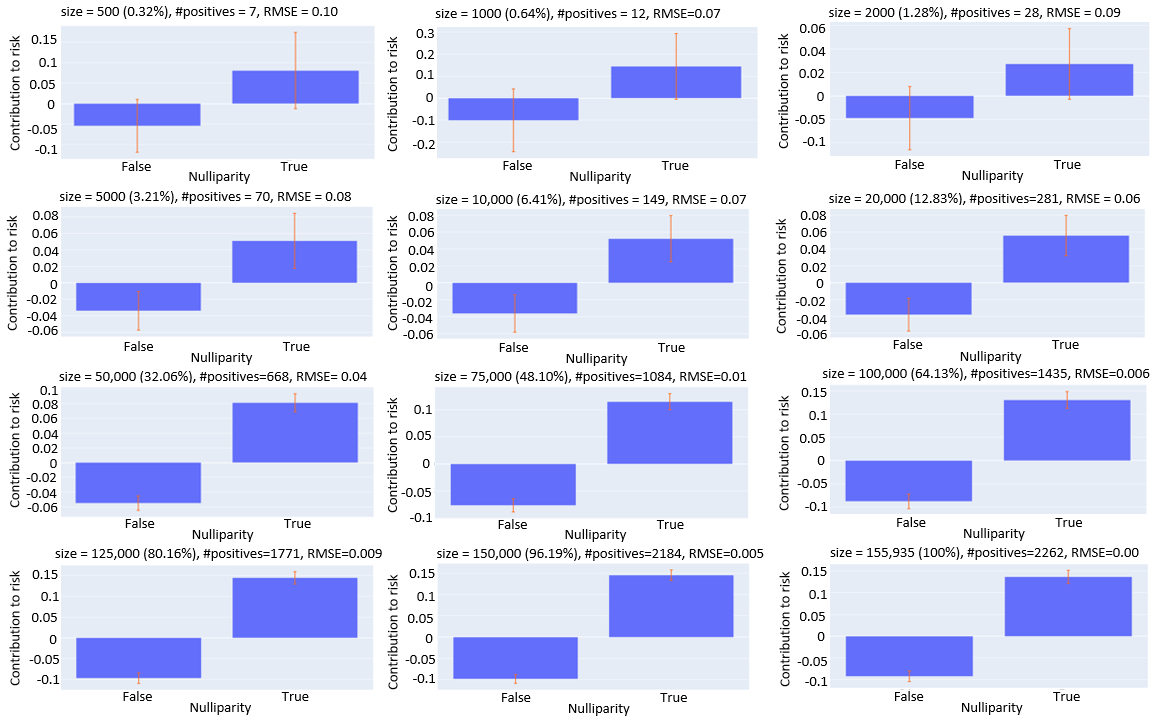}
    \caption{Evolution of the shape functions of \emph{nulliparity} while predicting SMM, as the data set size (``size" in the figure) is varied. We compute the RMSE by using the final shape function trained on all data as the reference.}
    \label{fig:SMM-shapefunc-evolution}
\end{figure}

Nulliparity is an important categorical feature that takes on either 0 or 1: 1 represents nulliparity is true, i.e. the mother has never given birth before, and 0 represents false nulliparity. The development of the shape function of nulliparity is visualized in Figure~\ref{fig:SMM-shapefunc-evolution}, showing the shape functions when trained on random subsets of the data set of sizes 500, 1000, 2000, 5000, 10000, 20000, 50000, 75000, 100000, 125000, 150000, and the entire data set of size 155935. Despite the number of positives in the sampled data set being small, the EBM appears to be able to pick up the signal for nulliparity even with small subsets of the data, presumably because nulliparity is Boolean.

\subsubsection{Antepartum stillbirth} \label{sec:robustness_SB}
\begin{figure}[!h]
    \centering
    \includegraphics[width=\linewidth]{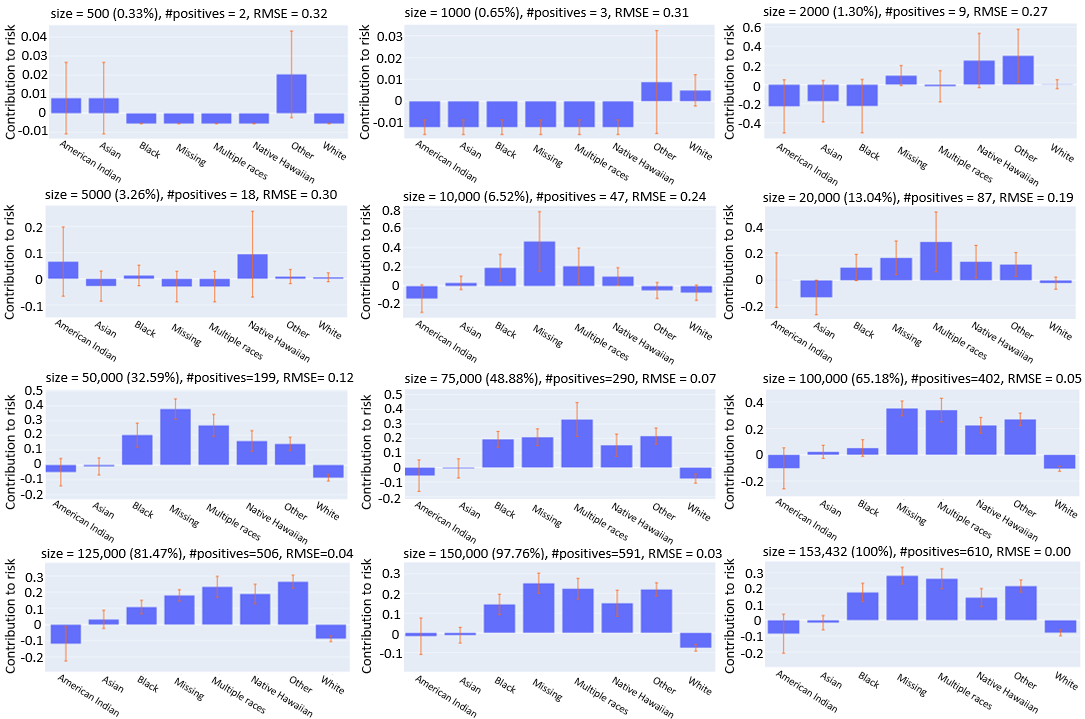}
    \caption{This shows the development of the EBM's shape function of \textit{Race} when predicting antepartum stillbirth as the data set size (``size" in the figure) is varied.}
    \label{fig:SB-evolution_shapefuncs}
\end{figure}

For antepartum stillbirth we pick \emph{race} as feature to explore, which is categorical with multiple classes, ranking second in the feature importance list, see Appendix~\ref{sec_App:ft_ranking_SB}. Similar to preterm preeclampsia, we select data set sizes of 500, 1000, 2000, 5000, 10000, 20000, 50000, 75000, 100000, 125000, 150000, and 153432 (complete data set). See Figure~\ref{fig:SB-evolution_shapefuncs}.

\subsubsection{Convergence of shape functions}
\begin{figure*}[!h]
    \centering
    \begin{subfigure}[t]{.48\textwidth}
        \centering
        \includegraphics[width=1\linewidth]{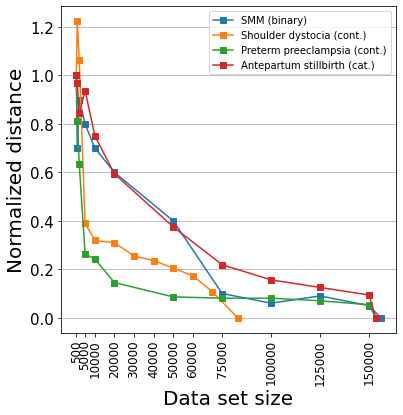}
        \caption{Sequence of normalized distances as a function of the data set size.}
        \label{fig:Frechet-datasetsize}
    \end{subfigure}%
    \hfill
    \begin{subfigure}[t]{.48\textwidth}
        \centering
        \includegraphics[width=1\linewidth]{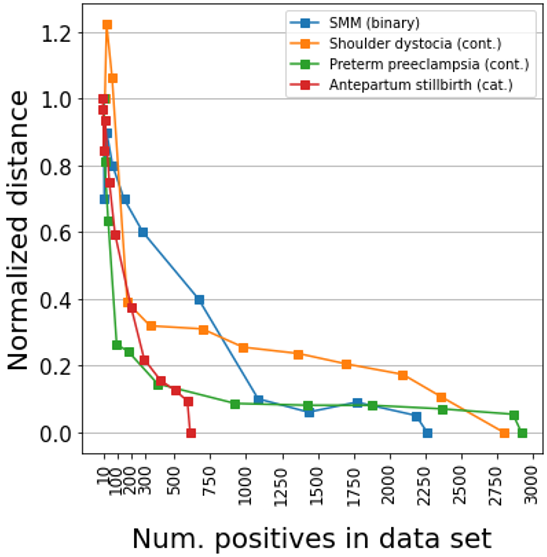}
        \caption{Sequence of normalized distances as a function of the number of positives in the sampled data set.}
        \label{fig:frechetdist_posDS}
    \end{subfigure}
    \caption{We plot the sequence of normalized distances where we consider the discrete Fr\'echet distance for continuous features (`cont.') and RMSE for categorical features (`cat.' or `binary'). By normalization we mean dividing the distances in each sequence by the first (usually largest) distance in the sequence, to better visualize how fast relative convergence occurs. By definition, the distances for the full sample size are zero.}
\label{fig:frechetdists}
\end{figure*}

We observe that the EBMs' shape functions convergence quickly as a function of the data set size and number of positives. For continuous features we compute the Fr\'echet distances, and denote these for each shape function trained on a subset of the full data set, see Figures~\ref{fig:ShDys-shapefunc-evolution} and \ref{fig:PP-shapefunc-evolution}. Because the scales of risk contributions in Figures~\ref{fig:ShDys-shapefunc-evolution} and \ref{fig:PP-shapefunc-evolution} are very different, so are the corresponding Fr\'echet distances. We normalize the Fr\'echet distances by dividing the distances by the first in order to more accurately show the relative convergence of the sequence of shape functions. See Figure~\ref{fig:frechetdists}.

For shape function evolutions of categorical features, as in Figures~\ref{fig:SMM-shapefunc-evolution} and \ref{fig:SB-evolution_shapefuncs}, the main impact of having a larger data set once there are enough samples to have positive and negative cases for each value seems to be variance reduction (smaller error bars). In each sequence of shape functions of categorical features, we compute the RMSE by using the final shape function trained on all data as the reference, and plot the normalized RMSEs in Figure~\ref{fig:frechetdists}.

\section{Discussion}\label{sec:discussion}
We perform external validation to confirm that the models generalize well to hospitals and settings different from the training set, and achieve good performance. In this external validation, we carefully control for hospital level of care to ensure fair representation of different subpopulations, to reduce potential bias in the model as much as possible, and to ensure the models attain good calibration on different demographic groups. Unfortunately, few models are externally validated, and failure to use external validation increases the risk that the model and results will not generalize across different populations. \cite{steyerberg2016prediction, wynants2017key, kleinrouweler2016prognostic}. This strengthens our confidence in the models and its findings. 

As shown in Table~\ref{tab:aurocs} and Figure~\ref{fig:cal-plots}, EBMs are well-calibrated models with AUROCs as good as other models. The excellent calibration of the EBM models suggests they perform well across the entire spectrum of patient risk. Additionally, a key distinguishing factor between EBMs and other ML models is the EBM's interpretability --- the shape functions $f_i(x_i)$ in the GAM can easily be visualized, see Section~\ref{sec:EBM}. Importantly, EBMs find that the largest contributors to risk are not always the traditionally recognized ones. We now examine each of the four outcomes in more detail.

\paragraph{Shoulder dystocia}
The main risk factors usually associated with shoulder dystocia are diabetes and baby birthweight. While the EBM reiterates the great influence of birthweight, it also shows that the time from admission to complete cervical dilation and the mother's height are very important.

Figure~\ref{fig:ShD-babyweight_and_maternal_height} uses actual birthweight (as opposed to estimated birthweight) to highlight the importance of fetal weight for shoulder dystocia. This suggests the value of more research and improved prenatal estimation of fetal weight in predicting maternal and fetal outcomes. Additionally, the importance of maternal height to the risk of shoulder dystocia suggests the potential value of more accurate measurement of pelvic dimensions.

\paragraph{SMM}
The main indicators picked up by the EBM for SMM are preeclampsia, labor type, nulliparity, initial cervical dilation, and race. Maternal age appears to be not as important of a predictor, even though it's traditionally recognized as an important factor; see Figure~\ref{fig:SMM-ft_imp} and Appendix~\ref{sec_App:ft_ranking_SMM}.  

\paragraph{Preterm preeclampsia}
The largest contributions to the risk of preterm preeclampsia are made by the mother's BMI, nulliparity and chronic hypertension. The mother's age also plays a major role, especially for women older than 44, where there is a steep increase in risk, see Figure~\ref{fig:PP-mothersAge_BMI}.

\paragraph{Antepartum stillbirth}
We find we can model antepartum stillbirth well, even in the face of a very low positive rate. Some of the most important features are BMI, alcohol use, and nicotine use, all of which are potentially modifiable for a subset of the population. This suggests steps could be taken to lower the average antepartum stillbirth risk nationwide. Other important features include race, age, height, and socioeconomic factors. Surprisingly, chronic hypertension was not one of the more significant factors in predicting antepartum stillbirth, perhaps reflecting greater surveillance and earlier timing of delivery for patients identified as having chronic hypertension.

\paragraph{Race and ethnicity}
We recognize that including race and ethnicity in clinical models is a complex issue. We propose one potential approach which is to include race and ethnicity in the model, let the interpretable model tell us what the statistics indicate about race and ethnicity, and then, if needed, mitigate bias by zeroing out the learned effects for these terms.




\section{Conclusion}\label{sec:conclusions}
We train a variety of models (logistic regression, EBMs, Random Forests, Boosted Trees and Neural Nets) for each of four important pregancy complications: SMM, shoulder dystocia, preterm preeclampsia and antepartum stillbirth.  Importantly, external validation is used to evaluate all of the models.  The results suggest that the interpretable EBM models have accuracy equal to or better than the other models; exhibit better calibration than the other models; and their interpretability allows us to discover surprising clinical effects that in some cases challenge traditional beliefs. An analysis of the convergence of the risk term profiles shows the sample sizes sufficient to train robust models.
A key advantage of interpretable learning methods such as EBMs is that they make it possible to go beyond predicting probabilities but also allow clinicians to understand what the model has learned about the clinical problem.
Some of the effects observed in the models may provide opportunities for interventions that have the potential to improve maternal and fetal outcomes.

\section*{Declarations}



\textbf{Competing Interests}: Vivienne Souter is a medical director at Natera. The authors declare no competing interests otherwise.







\begin{appendices}

\section{Lists of features}\label{sec_App:List_features_rankings}
For each of the four outcomes we provide a full list of features used. The features are ranked according to their feature importance from highest (first feature) to lowest (last feature) as computed by the EBM. Feature importance is measured as the mean absolute contribution to the log odds of the risk prediction over all samples.

\subsection{Severe maternal morbidity}\label{sec_App:ft_ranking_SMM}
(1) Preeclampsia/gestational hypertension, (2) labor type (spontaneous labor or induction of labor) , (3) nulliparity, (4) initial cervical dilation, (5) race, (6) Distressed Communities Index quintile (Economic Innovation Group), (7) Scheduled cesarean (planned cesarean birth), (8) initial maternal BMI (either pre-pregnancy or at first prenatal visit), (9) labor allowed (vaginal birth attempted), (10) hour of admission, (11) Hispanic/Latina ethnicity, (12) gravidity (the total number of pregnancies the pregnant person has had including the current pregnancy), (13) number of previous stillbirths, (14) gestational diabetes, (15) history of uterine surgery, (16) maternal age, (17) month of admission, (18)  IVF, (19) final maternal BMI, (20) cervical ripening, (21) maternal height, (22) history of classical incision, (23) insurance type, (24) absence minimal prenatal care, (25) number of previous cesareans, (26) induced labor indication, (27) history of low vertical incision, (28) parity, (29) membrane status (membranes ruptured or intact at the time of admission to labor and delivery), (30) presentation at delivery, (31) placental abruption, (32) history of uterine rupture, (33) day of week at admission, (34) sex of baby, (35) illicit substance use during pregnancy, (36) pre-pregnancy diagnosis of mental illness, (37) Rural-Urban Commuting Area (RUCA) code class, (38) pre-pregnancy diagnosis of diabetes, (39) ASA recommended for use during pregnancy (documentatoin of low dose aspirin recommendation in the medical record), (40) pre-pregnancy use of nicotine, (41) pre-pregnancy use of marijuana, (42) use of nicotine during pregnancy, (43) pre-pregnancy illicit substance use, (44) chronic hypertension, (45) cholestasis of pregnancy, (46) use of marijuana during pregnancy, (47) pre-pregnancy use of alcohol, (48) use of alcohol during pregnancy, (49) number of previous preterm births.

\subsection{Shoulder dystocia} \label{sec_App:ft_ranking_ShDys}
(1) Birthweight, (2) maternal height, (3) time from admission to complete cervical dilation, (4) final maternal BMI, (5) oxytocin use, (6) race, (7) Distressed Communities Index quintile, (8) Insurance type, (9) initial cervical dilation, (10) use of regional anesthesia, (11) time from complete dilation till delivery, (12) maternal age, (13) nulliparity, (14) number of weeks on delivery, (15) gravidity, (16) initial maternal BMI, (17) indication for operative vaginal delivery, (18) IVF, (19) cervical effacement, (20) month of admission, (21) membrane status, (22) vacuum use, (23) cervical ripening, (24) hour of admission, (25) pre-pregnancy diagnosis of diabetes, (26) pre-pregnancy diagnosis of mental illness, (27) gestational diabetes, (28) pre-pregnancy illicit substance use, (29) Hispanic/Latina Ethnicity, (30) Forceps, (31) day of week at admission, (32) Rural-Urban Commuting Area (RUCA) code class, (33) indication for induced labor, (34) pre-pregnancy use of alcohol, (35) history of uterine rupture, (36) nicotine use during pregnancy, (37) sex of baby, (38) pre-pregnancy use of marijuana, (39) alcohol use during pregnancy, (40) labor type, (41) parity (number of previous births of $\geq$ 20 weeks' gestation), (42) cholestasis of pregnancy, (43) preeclampsia/gestational hypertension, (44) use of marijuana during pregnancy, (45) use of illicit substance during pregnancy, (46) absent of minimal prenatal care, (47) ASA recommended for use during pregnancy, (48) pre-pregnancy use of nicotine, (49) number of previous preterm births, (50) number of previous stillbirths, (51) history of uterine surgical history, (52) history of classical incision, (53) pre-pregnancy diagnosis of hypertension, (54) history of low vertical incision, (55) number of previous cesarean births, (56) placental abruption.

\subsection{Preterm preeclampsia}\label{sec_App:ft_ranking_PP}
(1) BMI (2) nulliparity, (3) chronic hypertension, (4) maternal age, (5) pre-pregnancy diagnosis of mental illness, (6) number of previous stillbirths, (7) gravidity, (8) race, (9) Distressed Communities Index quintile, (10) number of previous preterm births, (11) Rural-Urban Commuting Area (RUCA) code class, (12) pre-pregnancy diagnosis of diabetes, (13) maternal height, (14) Hispanic/Latina ethnicity, (15) illicit substance use during pregnancy, (16) previous low vertical incision, (17) (other) uterine surgical history, (18) number of previous cesarean births, (19) history of uterine rupture, (20) marijuana use during pregnancy (21) pre-pregnancy use of marijuana, (22) pre-pregnancy illicit substance use, (23) IVF, (24) sex of baby, (25) pre-pregnancy use of alcohol, (26) previous classical cesarean, (27) alcohol use during pregnancy, (28) nicotine use during pregnancy, (29) pre-pregnancy use of nicotine.

\subsection{Antepartum stillbirth} \label{sec_App:ft_ranking_SB}
(1) BMI, (2) race, (3) Distressed Communities Index quintile, (4) maternal age, (5) maternal height, (6) Rural-Urban Commuting Area (RUCA) code class, (7) gravidity, (8) pre-pregnancy diagnosis of diabetes, (9) number of previous cesarean births, (10) number of previous stillbirths, (11) pre-pregnancy use of alcohol, (12) pre-pregnancy use of nicotine, (13) IVF, (14) baby gender, (15) nulliparity, (16) history of classical incision, (17) history of uterine surgical history, (18) previous low vertical incision, (19) pre-pregnancy use of marijuana, (20) chronic hypertension, (21) nicotine use in pregnancy, (22) pre-pregnancy diagnosis of mental illness, (23) marijuana use during pregnancy, (24) illicit substance use during pregnancy, (25) Hispanic/Latina ethnicity, (26) number of previous preterm births, (27) pre-pregnancy illicit substance use, (28) alcohol use during pregnancy, (29) history of uterine rupture.




\end{appendices}


\bibliography{sn-bibliography}


\end{document}